\documentclass{llncs}
\usepackage{makeidx}  
\usepackage{times}
\usepackage{latexsym}
\usepackage{url}
\usepackage{amsmath}
\usepackage{amssymb}
\usepackage{multirow}
\usepackage{graphicx}
\usepackage{caption}
\usepackage{verbatim}
\usepackage[numbers,sort]{natbib}
\usepackage{enumitem}
\usepackage{color}
\usepackage[justification=centering]{caption}
\usepackage{courier}

\title{Using Word Embeddings for Query Translation for Hindi to English Cross Language Information Retrieval}

\author{Paheli Bhattacharya, Pawan Goyal
\and Sudeshna Sarkar}
\institute{Department of Computer Science and Engineering\\Indian Institute of Technology Kharagpur\\Kharagpur, West Bengal, India - 721302\\
  {\texttt{paheli@iitkgp.ac.in}}, \{\texttt{pawang,sudeshna\}@cse.iitkgp.ernet.in}}
\begin{document}
\frontmatter          
\pagestyle{headings}  

\maketitle

\begin{abstract}
Cross-Language Information Retrieval (CLIR) has become an important problem to solve in the recent years due to the growth of content in multiple languages in the Web.
One of the standard methods is to use query translation from source to target language. In this paper, we propose an approach based on word embeddings, a method that captures contextual clues for a particular word in the source language and gives those words as translations that occur in a similar context in the target language. 
Once we obtain the word embeddings of the source and target language pairs, we learn a projection from source to target word embeddings, making use of a dictionary with word translation pairs. We then propose various methods of query translation and aggregation. The advantage of this approach is that it does not require the corpora to be aligned (which is difficult to obtain for resource-scarce languages), a dictionary with word translation pairs is enough to train the word vectors for translation. 

We experiment with Forum for Information Retrieval and Evaluation (FIRE) 2008 and 2012 datasets for Hindi to English CLIR.
The proposed word embedding based approach outperforms the basic dictionary based approach by 70$\%$ and when the word embeddings are combined with the dictionary, the hybrid approach beats the baseline dictionary based method by 77$\%$. It outperforms the English monolingual baseline by 15$\%$, when combined with the translations obtained from Google Translate and Dictionary. 
\end{abstract}

\section{Introduction}
\label{intro}
English has been a dominating language of the Web for long but with the rising popularity of the Web, native languages have also found their places - now the Web has substantial content in multiple languages. This prompted the task of Cross Language Information Retrieval (CLIR), where the language of the documents being queried is different from the query language.
One of the main motivations behind CLIR is to gather a lot of knowledge from a variety of knowledge bases which are in the form of documents in various languages, helping a diverse set of users, who can provide the queries in the language of their choice.

Intuitively, Cross Language Information Retrieval is harder than Monolingual Information Retrieval because it needs to cross the language boundaries either by translating the query or by translating the document or by translating both the query and the document to a third language. 
There are many techniques to implement CLIR. One way to translate the query is a token-to-token translation based approach that uses a machine readable dictionary \cite{dict:1,dict:2,dict:3}. Another is to employ Statistical Machine Translation (SMT) systems \cite{smt-coling:12,smt-sigir:12,smt-sigir:14} to translate the query. SMT is a machine translation technique that leverages statistical models whose parameters are derived using parallel bilingual corpora. Other methods for query translation include corpus based techniques \cite{cicling:15}, using online translation services like Google Translate \cite{www:15} or by using large scale multilingual resources like Wikipedia \cite{ecir:11}. 
 
Most of these approaches require either a full fledged dictionary, an aligned corpora or a machine translation system, which may not be guaranteed for resource scarce languages. In this paper, we attempt to solve the problem in a scenario when the monolingual corpus is available in both the languages, but may not be aligned. Additionally, a few word pair translations between the two languages are required, but these need not be exhaustive. We study the effectiveness of word embeddings based methods in this scenario.

In word embeddings, words from the vocabulary are mapped to vectors of real numbers in a low dimensional space; and these vectors are called as embeddings. It has been seen that in the distributed space defined by the vector dimensions, 
syntactically and semantically similar words fall closer to each other. 
Given a training corpus, word embeddings are able to generalize well over words that occur less frequently as well.
In this paper we try to explore how the usage of word embeddings can affect the retrieval performance in a CLIR based system. 
To the best of our knowledge, no such approach using comparable corpora has been tried out for the CLIR tasks. 

Handling Out-Of-Vocabulary (OOV) terms that are not named entities is a major technical difficulty in CLIR task. For Hindi words that are actually part of the English vocabulary, for example, {`\sl kaiMsara'}\footnote{All Hindi words have been written in ITrans using \url{http://sanskritlibrary.org/transcodeText.html}} (meaning, cancer), {\sl `aspataala'} (meaning, hospital), dictionary and corpus based methods had to resort to ``transliteration", but the embedding based method captured their contextual cues and was able to find related words in English. Words brought out as translations for {\sl `kaiMsara'} were `cancer',`disease',`leukemia', for {\sl `aspataala'} the words that came out as translations were `hospital',`doctor',`ambulance'. We perform transliterations only to handle the named entities.

We also propose and compare various techniques for aggregating the target translations using multiple query terms. We find that instead of aggregating the query vector at the source side, if we compute the similarity scores for each query term separately and then aggregate the resulting vectors, it provides better performance. Our proposed word embedding based approach and the hybrid approach (combined with dictionary) could achieve 88$\%$ and 92$\%$ of the Mean Average Precision (MAP) as reported by the English monolingual baseline, respectively. When combined with translations obtained from Google Translate, it was able to beat the English monolingual MAP by 15\%. The methods also showed improvements of 29$\%$, 34$\%$ and 68$\%$ over \cite{clef:07}, a state-of-the-art corpus based approach.
\section{Related Work}
\subsection{Cross-Language Information Retrieval}
\label{rw-clir}
People have tried viewing Cross-Language Information Retrieval (CLIR) from various aspects. To start with, \cite{dict:2} uses dictionary based translation techniques for Information Retrieval. They use two dictionaries, one, in which general translation of a query term is present and the other, in which, domain-specific translation of the query term is present.
\cite{dict-new:05} discusses the key issues in dictionary-based CLIR. They have shown that query expansion effects are sensitive to the presence of orthographic cognates and develop a unified framework for term selection and term translation.
\cite{lsi:1,lsi:2} perform CLIR by computing Latent Semantic Indexing on the term-document matrix obtained from a parallel corpora. After reducing the rank, the queries and the documents are projected to a lower dimensional space.

Statistical Machine Translation (SMT) techniques and its improvements have also been tried out \cite{smt-acl:14,smt-coling:12,smt-sigir:14}. \cite{clef2:07} uses SMT for CLIR between Indian languages. They use a word alignment table that was learnt using an SMT on parallel sentences to translate source language query to target language query. In \cite{smt-sigir:14}, the SMT technique was trained to produce a weighted list of alternatives for query translation. 

Transliteration based models have also been looked into. \cite{ecir:09} uses transliteration of the Out-Of-Vocabulary (OOV) terms. They treat a query and a document as comparable and for each word in the query and each word in the document, they find out a transliteration similarity value. If this value is above a particular threshold, then the word is treated as a translation of the source query word. They iterate through this process, working on relevant documents retrieved in each iteration. \cite{clef:07} uses a simple rule based transliteration approach for converting OOV Hindi terms to English and then uses a pageRank based algorithm to resolve between multiple dictionary-translations and transliterations. 


\cite{ecir:11} uses Wikipedia concepts along with Google translate to translate queries. The Wikipedia concepts are mined using cross-language links and redirects and a translation table is built. Translations from Google are then expanded using these concept mappings. Explicit Semantic Analysis (ESA) is a method to represent documents in the Wikipedia article space as vectors whose components represent its association with the Wikipedia articles. \cite{esa:08} uses it in CLIR along with a mapping function that uses cross-lingual links to link documents in the two languages that talk about the same topic. Both the queries and the documents are mapped to this ESA space, where the retrieval is performed. 

\cite{babelnet} leverages BabelNet, a multilingual semantic network. They build a basic vector represenation of each term in a document and a knowledge graph for every document using BabelNet and interpolate them in order to find the knowledge-based document similarity measure.

Similarity Learning via Siamese Neural Network \cite{s2net} trains two identical networks concurrently in which the input layer corresponds to the original term vector and the output layer is the projected concept vector. The model is trained by minimizing the loss of the similarity scores of the output vectors, given pairs of raw term-vectors and their labels (similar or not).

 \cite{www:15} uses online translation services, Google and Bing, to translate queries from source language to target language. They conclude that no single perfect SMT or online translation service exists, but for each query one performs better than the others.
 
\subsection{Word Embedding}
\label{rw-we}
\cite{wordvecorig:13} proposed a neural architecture that learns word representations by predicting neighbouring words. There are two main methods by which the distributed word representations can be learnt. One is the Continuous Bag-of-Words (CBOW) model that combines the representations of the surrounding words to predict the word in the middle. The second is the Skip-gram model that predicts the context of the target word in the same sentence. GloVe or Global Vectors \cite{glove:14} is also an unsupervised algorithm for learning word representations. The training objective of GloVe is to learn word vectors such that for any pair, the dot product equals the log of the words' probability of co-occurrence. They use global matrix factorization and local context window methods to build global vectors.

Word embedding based methods have been utilized in many different tasks, such as word similarity \cite{manaal,morpheme2,partha:15}, cross lingual dependency parsing \cite{partha:15}, finding semantic and syntactic relations \cite{manaal}, finding morphological tags \cite{morpheme1}, identifying POS and translation equivalence classes \cite{taskspecific} and in analogical reasoning \cite{morpheme2}. \cite{wordvec:13} uses the word vectors to translate between languages. Once the word vectors of the two languages have been obtained, it builds a translation matrix using stochastic gradient descent version of linear regression that transforms the source language word vectors to the target language space.


\subsection{Word Embedding based CLIR}
\label{rw-weclir}
\cite{ivan:15} leverages document aligned bilingual corpora for learning embeddings of words from both the languages. Given a document \textit{d} in a source language and its comparable document aligned equivalent \textit{t} in the target language, they merge and randomly shuffle the documents \textit{d} and \textit{t}. They train this ``pseudo-bilingual" document using word2vec. To get the document and query representations, they treat them as bag-of-words and combine the vectors of each word to obtain the representations of query and document. Between a query vector and a document vector, they compute the cosine similarity score and rank the documents according to this metric.


In this paper, we attempt to perform CLIR from Hindi to English using translations obtained from word embedding based methods. The main advantage of word embeddings is that it does not suffer from data sparsity problems. Given a training corpus, they are able to generalize well over words that occur less frequently. Additionally, they are also computationally efficient \cite{wordvecorig:13}.



\section{The Proposed Framework}
We use the query translation approach towards Hindi to English CLIR, that is, we translate Hindi queries to English and perform monolingual information retrieval on English documents. Towards query translation, we first obtain word embeddings for both the source and target languages using corpus for individual languages. Then, we learn a projection function from source to target word embeddings using aligned word pairs, as obtained from the dictionary. Finally, we employ various methods for query translations: one in which every query term in the source language has \textit{k} best translations in the target language. The second, in which we aggregate the query word vectors into a single vector that represents the query as a whole and then obtain \textit{k} best translations for the query itself.

\subsection{Dataset}
\label{dataset}
We have used the FIRE (Forum for Information Retrieval Evaluation, developed as a South-Asian counterpart of CLEF, TREC, NTCIR) 2012 and 2008 datasets obtained from \footnote{\url{http://fire.irsi.res.in/fire/data}}. The FIRE 2012 corpus contains 392,577 English documents (from the newspapers -- `The Telegraph' and `BDNews 24') and 367,429 Hindi documents (from the newspapers -- `Amar Ujala' and `Navbharat Times'). For FIRE 2008, we used the same number of English documents\footnote{We could not get the actual English documents for 2008 after repeated trials, so we used the updated dataset of 2012. The actual dataset was a subset of 2012 dataset.} and 95,215 Hindi documents (from the Hindi newspaper `Dainik Jagran'). The corpora are comparable but not aligned. The queries for the CLIR task of FIRE were ranging from topics 176-225 and 26-75 for 2012 and 2008, respectively. We use the title field for the experiments. The English-Hindi dictionary is obtained from \url{http://ltrc.iiit.ac.in/onlineServices/Dictionaries/Dict_Frame.html}. It also contains translations that were multi-word. We exclude these translation pairs for our experiments. We obtain the stopword list from \url{http://www.ranks.nl/stopwords/hindi} and English Named-Entity Recognizer from \url{http://nlp.stanford.edu/software/CRF-NER.shtml}

Next, we discuss in detail various steps in our framework.

\subsection{Obtaining Word Embeddings for the Source and Target Languages}
\label{embeddings}
We use word2vec introduced by \cite{wordvecorig:13}. We train the word2vec package\footnote{Obtained from \url{https://code.google.com/p/word2vec/}} for both the monolingual datasets of English and Hindi. We use the CBOW model with a window size of 5 and output vector of 200 dimensions with other default parameters set.

\subsection{Learning the Projection of Word Embeddings from the Source to the Target Language Space}
\label{embed-trans}
We use linear regression to learn a projection from the source to the target language space, similar to an approach used by \cite{wordvec:13}.
The idea is as follows: Given a translation dictionary, we extract the word embeddings of the translation pair $\left \{x{_i},y{_i}\right \}$ where {$x_i$} $\in$ $\mathbb{R}$\textsuperscript{d$_1$} is a $d_1$- dimensional embedding learnt from the Hindi corpus for $x_i$ and y{$_i$} $\in$ $\mathbb{R}$\textsuperscript{d$_2$} is a $d_2$- dimensional embedding learnt from the English corpus for $y_i$. The aim is to find a translation matrix $W$ from the source to target such that the root mean square error between $Wx_i$ and $y_i$ is minimized. 

After obtaining the translation matrix $W$ using linear regression, embeddings for each word in Hindi (${w_h}$) can be multiplied with $W$ to obtain the equivalent vector $v$ of w${_h}$ in the target language space ($v=Ww_h$).
\subsection{Query Translation Process}
\label{query trans}

Given a query Q and its terms $q_1, q_2,\ldots,q_n$, we first remove the stop-words from the query. We then use the vector space embedding of each query term {$q_i$}, along with the embeddings of all the English words, as obtained using the embedding based method described in Section \ref{embeddings}, to translate this query, while making use of the translation matrix, obtained in Section \ref{embed-trans}. We adopt the following methods for query translation:

\begin{itemize}[leftmargin=*]


\item \textbf{Word embedding (WE) to translate each query term independently}:  
In this approach, once we get the word vector of each query term projected in the target language ($v$), we compute the cosine similarity between the vector embedding of each English word and $v$ and pick the \textit{k} best translations for this query term. 
An example of a query and its 3 best translations is as follows:\\
\textbf{Query in Hindi:} {\sl 2008 guvaahaaTii bama visphoTa se xati} \\
\textbf{Meaning in English:} Loss due to 2008 Guwahati explosions\\
The translations of the query terms are given in Table \ref{table:eg-query-translation}. {\sl 2008} and {\sl guvaahaaTii} are treated as Named Entities (details in Section \ref{transliteration}) and hence have one translation each. We see that the WE method gives related words for each query term. We add the translations obtained independently from each query term to obtain the final translation but each term is weighted uniformly.

\begin{table*}
\vspace{-0.6cm}
    \centering
    \caption{Translations of query terms for ``{\sl 2008 guvaahaaTii bama visphoTa se xati}" using WE}
    \label{table:eg-query-translation}
         \begin{tabular}{|c|c|c|}
    \hline
    \centering
     \textbf{Query Term in Hindi} & \textbf{Meaning in English} & \textbf{Translations using Embeddings}\\
     \hline
     2008 & 2008, year & 2008\\
     \hline
     {\sl guvaahaaTii} & Guwahati, a place in India & Guwahati\\
     \hline
     { \sl bama} & bomb & explosives, bomb, device\\
     \hline
     {\sl visphoTa} & explosion & explosion, blast, accident\\
     \hline
     {\sl xati} & loss & degradation, damage, distortion\\
     \hline
    \end{tabular}
    \vspace{-0.4cm}
\end{table*}
\item \textbf{WE weighted}: Assigning weights to query words is necessary to distinguish between words that are important in a query from words that are not. 
In this approach, we proportionally distribute the weights according to the similarity score for each translated word with the query word(s). We then normalize the translated query so that the weights for all translations terms add up to 1. 

\item \textbf{Combining Similarity Vectors for Translations (SIM Vec)}: In this approach, instead of treating each query term independently, we aggregate the results by combining results from each query term. One possible way is to combine the vector components at the source\footnote{We have tried the sum, max and min combinations, but they do not give good result.}. Instead, we first map each query term to the target space, then compute similarity values for each query term with the target words, and combine these similarity values. Thus, for a query word $q_j$, we build a vector $V_j$, where the {$i^{th}$} component of the vector, $V_j[i]$, denotes the similarity value of that particular word with the {$i^{th}$} target language word in the vocabulary. 
Suppose there are 5 words in the English vocabulary - cricket, football, game, laptop and computer and suppose we want to build the similarity vector of the Hindi word {\sl khela}. The cosine similarity values are listed in Table 2. The similarity vector of {\sl khela} can be written as: $\left [ 0.64 \ 0.69\  0.8\  0.32\  0.25 \right ]$ \\
\begin{table*}
\vspace{-0.6cm}
\centering
\caption{Example to illustrate SIM Vec. The table shows Cosine Similarity Values between the Hindi word \textit{khela} (which means `game') with other English words.}
\label{my-label}
\begin{tabular}{|c|c|c|}
\hline
Word in Hindi & Word in English & \begin{tabular}[c]{@{}c@{}}Cosine Similarity \\ Value\end{tabular} \\ \hline
\multirow{5}{*}{{\sl khela}} & cricket & 0.64 \\ \cline{2-3} 
 & football & 0.69 \\ \cline{2-3} 
 & game & 0.8 \\ \cline{2-3} 
 & laptop & 0.32 \\ \cline{2-3} 
 & computer & 0.25 \\ \hline
\end{tabular}
\vspace{-0.4cm}
\end{table*}


Now, once we obtain such vectors for each query term, these vector components are merged using the summation or the maximum function. The idea behind using the `summation' function is to find which words in the target language (English) vocabulary is the most similar when there is a contribution by all the source language query terms. The `maximum' function provides knowledge as to which word in the target language vocabulary is maximally correlated to any of the source language query terms. The formula for finding the resultant query vector ($V_{sum}$ and $V_{max}$, for the `summation' and `maximum' functions, respectively) from the vectors of the similarity values are shown in Equations \ref{eqn:sum} and \ref{eqn:max}. $n$ denotes the number of terms in the query and $d$ denotes the number of words in the target language vocabulary.

\begin{eqnarray}
\vspace{-0.8cm}
\label{eqn:sum}
V_{s · u · m}[i] =\sum_{j=1}^{n} V_{j}\left [ i \right ]\\
\begin{aligned}
\label{eqn:max}
V _{m · a · x}[i] = \max_j \left ( V_{j}\left [ i \right ] \right ) \\
\forall j, 1\leqslant j \leqslant n ; 
\forall i, 1 \leqslant i \leqslant d  
\end{aligned}
\end{eqnarray}
From the resultant vector, we extract the top \textit{k} target language vocabulary words with the highest scores.

\end{itemize}
\subsection{Transliteration of Named Entities}
\label{transliteration}
The source language query also contains named entities, which may not be present in the vocabulary. Since no Named-Entity Recognition (NER) tool is available for Hindi, we resort to the transliteration based process. For each Hindi character, we construct a table of its possible transliterations. For example, the first consonant in Hindi {\sl ka} has 3 possible transliterations in English -- {\sl ka, qa, ca}. We apply several language specific rules - a consonant, for instance {\sl ka} in Hindi can have two forms, one that is succeeded by a silent {\sl a}, i.e., {\sl ka} and another that is not, i.e., {\sl k}. The second case applies when it is succeeded by a vowel or another consonant in conjunction (also known as {\sl yuktakshar}). For each transliteration of an OOV Hindi query word  $h$ and for each word $e$ in the list of words returned as named entities in English language, we apply the Minimum Edit Distance algorithm between $h$ and $e$. We then take the word with the least edit distance. Our transliteration concept is based on \cite{clef:07} and gives quite a satisfactory result, with an accuracy of 90$\%$.
\section{Experiments}
We used Apache Solr version 4.1 as the monolingual retrieval engine. The similarity score for the query and the documents was the default TF-IDF Similarity\footnote{\url{https://lucene.apache.org/core/3_5_0/api/core/org/apache/lucene/search/Similarity.html}}.
The human relevance judgments were available from FIRE. Each query had about 500 documents that were manually judged as relevant (1) or non-relevant (0). We then used the \textit{trec-eval} tool \footnote{\url{http://trec.nist.gov/trec_eval/}} for finding the Precision at 5 and 10 (P5 and P10) and the Mean Average Precision (MAP).
\subsection{Baselines}
We use the following baselines for comparison. \textbf{English Monolingual} corresponds to the retrieval performance of the target language (English) queries supplied by FIRE. \textbf{Dictionary} is the dictionary based method where the query translations have been obtained from the dictionary. For words that contain multiple translations, we include all of them. Translations with multi-words are not considered. Named entities are handled as described in Section \ref{transliteration}. We also use the method proposed by \textbf{Chinnakotla et.al \cite{clef:07}} as a baseline since they participated in the FIRE task \cite{fire:08}\footnote{\cite{clef:07} is an improved version of \cite{fire:08}}. Finally, \textbf{Google Translate} is also used as a baseline, where the Hindi query is translated using Google Translate to English.

Results for these baselines are reported in Table \ref{tab:baseline}. \cite{clef:07} shows improvements over the dictionary since the OOV terms are transliterated and multiple dictionary translations are disambiguated using the contextual cues from the corpus, however it is not able to perform better than the monolingual baseline. \textbf{Google Translate}\footnote{\url{https://translate.google.com/}} outperforms the monolingual baselines.

\begin{table}
\vspace{-0.6cm}
\small
    \centering
    \caption{Performance Results for the Baseline approaches}
    \label{tab:baseline}
    \noindent\begin{tabular}{|c|c|c|c|c|c|c|c|c|c|}
    \hline
         & \multicolumn{3}{c|}{\textbf{2012 Dataset}} & \multicolumn{3}{c|}{\textbf{2008 Dataset}}  \\
         \cline{2-7}
         \textbf{Method} & \textbf{MAP} & \textbf{P5} & \textbf{P10} & \textbf{MAP} & \textbf{P5} & \textbf{P10}\\
         \hline
         English Monolingual &  0.3218 & 0.56 & 0.522 & 0.1609 & 0.248 & 0.236\\
         \hline
         Dictionary	&  0.1691 & 0.2048	& 0.2048 & 0.084 &	0.1464	& 0.137\\
         \hline
          Chinnakotla et.al \cite{clef:07} & 0.2236 & 0.3347 & 0.3388 & 0.11 & 0.15 & 0.147 \\
         \hline
         Google Translate & 0.3566 & 0.576 & 0.522 & 0.178 &	0.255 &	0.24\\
         \hline
          \end{tabular}
    
    \vspace{-0.1cm}
\end{table}

\subsection{Proposed Word embeddings based approaches}
\begin{table}[!thb]
\small
    \centering
    \caption{Performance Results when Queries are translated using proposed Word Embedding based methods: for WE and WE weighted, \# Translations per query term are shown, while for SIM Vec, \# Translations for the complete query are shown.}
    \label{tab:simple-we}
    \noindent\begin{tabular}{|c|c|c|c|c|c|c|c|}
    \hline
         & & \multicolumn{3}{c|}{\textbf{2012 Dataset}} & \multicolumn{3}{c|}{\textbf{2008 Dataset}}  \\
         \cline{3-8}
         \textbf{Method} & \textbf{\# Translations} & \textbf{MAP} & \textbf{P5} & \textbf{P10} & \textbf{MAP} & \textbf{P5} & \textbf{P10}\\
         \hline
         \multirow{3}{*}{WE} & 1word & 0.2533 & \textbf{0.3920} & \textbf{0.3840} & 0.1284 & \textbf{0.175} & \textbf{0.163}\\
         & 2words & \textbf{0.2568} & 0.3840 & 0.3720 & \textbf{0.129} & 0.167 & 0.154 \\
         & 3words & 0.2379 & 0.384 & 0.3520 & 0.127 & 	0.166 &	0.152\\
         & 5words & 0.2053 &	0.328 &	0.32 &	0.119 & 	0.145 &	0.143\\
         \hline
         \multirow{3}{*}{WE weighted} & 3words &	{0.2802} & \textbf{0.436} & {0.392} & {0.138} & {0.191} &	{0.187}\\
         & 5words & \textbf{0.2808} &	0.408 &	\textbf{0.408} &	\textbf{0.14} &	\textbf{0.218} &	\textbf{0.209}\\
         & 7words & 0.2804 &	0.428 &	0.402 &	0.136 &	0.21 &	0.196\\
         \hline
         \multirow{7}{*}{SIM Vec}
         & Sum - 15words & 0.2508 & 0.364 & 0.362 & 0.1276 & \textbf{0.2137} & \textbf{0.1968}\\
         & Sum - 20words & \textbf{0.2562} & \textbf{0.368} & \textbf{0.368} & \textbf{0.1282} & 0.2108 & 0.196\\
         & Sum - 25words & 0.2493 & 0.359 & 0.343 & 0.1268 & 0.187 & 0.1823 \\
         \cline{2-8}
         & Max - 10words & 0.2733 & 0.4120 & 0.382 & 0.138 & 0.23 & 0.225\\
         & Max - 15words & \textbf{0.2835} & 0.408 & 0.4 & \textbf{0.144} & 0.2416 & 0.237\\
         & Max - 20words & 0.2830 & 0.4120 & 0.392 & 0.14 & \textbf{0.2471} & 0.238\\
         & Max - 25words & 0.2812 & \textbf{0.424} & \textbf{0.394} & 0.137 & 0.24 & \textbf{0.24}\\
         \hline
           \end{tabular}
    
    \vspace{-0.1cm}
\end{table}

Table \ref{tab:simple-we} shows the performance of the proposed word embedding based approaches for query translation. Among the proposed approaches, \textbf{SIM Vec (max)} seems to perform the best on both the datasets. An issue that comes up while using the embedding based methods is whether to include the embeddings of the named entities in the process. 
For a particular word in the source language \textit{w}, similar words that showed up are relevant to \textit{w} but are
not translations. For example, the word \textit{BJP} in Hindi (which is an Indian political party) the words that were most similar also included the names of other political parties like \textit{Congress} and also words like \textit{Parliament} and \textit{government} in the target language English. Inclusion of such terms can harm the retrieval process as named entities play a critical role in Information Retrieval and so we decide to exclude them from the embeddings and use a transliteration scheme as described in Section \ref{transliteration}\\

\begin{table}[!thb]
\vspace{-0.6cm}
\small
\centering
\caption{Example queries which could not find Translations in the Dictionary but could find Translations using the proposed WE method}
\label{tab:eg-queries-we}
\noindent\begin{tabular}{|c|c|c|c|c|c|}
\hline
\textbf{Query in Hindi} & \textbf{Translation in English} & \textbf{Translations (WE)} & \textbf{MAP} & \textbf{P5} & \textbf{P10} \\ \hline
\begin{tabular}[c]{@{}c@{}}{\sl aadarsha haausiMga} \\ {\sl sosaaiTii}\\ {\sl ghoTaale istiiphaa}\end{tabular} & \begin{tabular}[c]{@{}c@{}}Adarsh Housing Society\\ scam resignation\end{tabular} & \begin{tabular}[c]{@{}c@{}}Adarsh housing \\ institution\\ scam coterie\end{tabular} & ~ 0.3 ~ & ~ 0.6 ~ & ~ 0.4 ~ \\ \hline
\begin{tabular}[c]{@{}c@{}}{\sl bhaaratiiya saMsada}\\ {\sl aataMkavaadii} \\ {\sl hamalaa}\end{tabular} & Indian Parliament attack & \begin{tabular}[c]{@{}c@{}}Indian Parliament \\ constitutional \\ terrorist assault\end{tabular} & ~ 0.21 ~ & ~ 0.6 ~ & ~ 0.6 ~ \\ \hline
\begin{tabular}[c]{@{}c@{}}{\sl aaiiphona aaiipaiDa} \\ {\sl Dijaaina}\\ {\sl lokapriyataa} \\ {\sl lancha}\end{tabular} & \begin{tabular}[c]{@{}c@{}}Design Popularity\\ iPhone iPad Launch\end{tabular} & \begin{tabular}[c]{@{}c@{}}iPhone iPad popularity\\ unveiled\end{tabular} & ~ 0.65 ~ & ~ 1 ~ & ~ 1~ \\ \hline
\end{tabular}
\vspace{-0.3cm}
\end{table}
\begin{table}[!thb]
\vspace{-0.1cm}
\small
\centering
\caption{Example queries to illustrate the `Max' and `Sum' functions for SIM Vec}
\label{tab:eg-queries-sum-max}
\resizebox{\textwidth}{!}{%
\noindent\begin{tabular}{|c|c|c|c|c|c|c|}
\hline
\textbf{\begin{tabular}[c]{@{}c@{}}Query \\ in Hindi\end{tabular}} & \textbf{\begin{tabular}[c]{@{}c@{}}Translation\\ in English\end{tabular}} & \textbf{\begin{tabular}[c]{@{}c@{}}Translation\\ Method\end{tabular}} & \textbf{ Translations } & \textbf{MAP} & \textbf{P5} & \textbf{P10} \\
\hline
\multirow{5}{*}{\begin{tabular}[c]{@{}c@{}}{\sl shriilaMkaaii} \\ {\sl raaShTriiya} \\ {\sl krikeTa} \\ {\sl Tiima para}\\ {\sl hamalaa}\end{tabular}} & \multirow{5}{*}{\begin{tabular}[c]{@{}c@{}}Sri Lankan\\ national\\ cricket \\ team attack\end{tabular}} &  Sum  & \begin{tabular}[c]{@{}c@{}}Sri\textasciicircum 1 Lankan\textasciicircum 1 \\ cricket\textasciicircum 0.34 team\textasciicircum 0.34\\ sport\textasciicircum 0.32\end{tabular} & ~ 0.3738 ~ & ~ 0.6 ~ & ~ 0.6 ~ \\
\cline{3-7}
 &  & \multirow{4}{*}{Max} & \multirow{4}{*}{\begin{tabular}[c]{@{}c@{}}Sri\textasciicircum 1 Lankan\textasciicircum 1 \\ team\textasciicircum 0.35 assault\textasciicircum 0.33 \\ attack\textasciicircum 0.32\end{tabular}} & \multirow{4}{*}{~ 0.51 ~} & \multirow{4}{*}{~ 0.8 ~} & \multirow{4}{*}{~ 0.9 ~} \\
 &  &  &  &  &  &  \\
 &  &  &  &  &  &  \\
 &  &  &  &  &  &  \\
 \hline
\multirow{2}{*}{\begin{tabular}[c]{@{}c@{}}{\sl iraaka kaa} \\ {\sl prathama}\\ {\sl chunaava}\end{tabular}} & \multirow{2}{*}{\begin{tabular}[c]{@{}c@{}}Iraq`s first\\ election\end{tabular}} &  Sum  & \begin{tabular}[c]{@{}c@{}}Iraqi\textasciicircum 1 choice\textasciicircum 0.37\\ unfashionable\textasciicircum 0.32 \\ predictable\textasciicircum 0.31\end{tabular} & ~ 0.08 ~ & ~ 0 ~ & ~ 0 ~ \\
\cline{3-7}
 &  &Max& \begin{tabular}[c]{@{}c@{}}Iraqi\textasciicircum 1\\ elections \textasciicircum 0.334 first\textasciicircum 0.332\\ election\textasciicircum 0.33\end{tabular} & ~ 0.4 ~ & ~ 0.8 ~ & ~ 0.6 ~ \\
 \hline
\multirow{2}{*}{\begin{tabular}[c]{@{}c@{}}{\sl miga} \\ {\sl durghaTanaa}\\ {\sl pashchima} \\ {\sl baMgaala}\end{tabular}} & \multirow{2}{*}{\begin{tabular}[c]{@{}c@{}}MiG crash \\ in West\\ Bengal\end{tabular}} & Sum  & \begin{tabular}[c]{@{}c@{}}MiG\textasciicircum 1 West \textasciicircum 1 \\ Bengal \textasciicircum 1 oriental\textasciicircum 0.34\\ venomous \textasciicircum 0.33 \\ exotic \textasciicircum 0.33\end{tabular} & ~ 0.18 ~ & ~ 0.2 ~ & ~ 0.2 ~ \\
\cline{3-7}
 &  &Max& \begin{tabular}[c]{@{}c@{}}MiG \textasciicircum 1 West \textasciicircum 1\\ Bengal \textasciicircum 1 accident \textasciicircum 0.36\\ mishap \textasciicircum 0.33 crash \textasciicircum 0.31\end{tabular} & ~ 0.4 ~ & ~ 0.8 ~ & ~ 0.5 ~
 \\
 \hline
\end{tabular}
}

\vspace{-0.6cm}
\end{table}
On further investigation, we find that there are 8 such queries for which no translation was available from the Dictionary. Table \ref{tab:eg-queries-we} shows some of these queries. For OOV words that are actually in English and have been written in Hindi orthographic format (e.g, `housing', `speaker' and `cancer' in English have been written as `{\sl haausiMga}', `{\sl spiikara}' and `{\sl kaiMsara}' in Hindi), word embeddings (WE) can easily retrieve translations like `housing',`society' and `speaker',`parliament'  and `cancer',`disease' respectively using contextual cues. It is thus evident that the word embedding based method is robust, the translations being very close in meaning to the source language words.

When weights are assigned to the translated words, the performance is even better.  The insight gained after observing the individual query results for the weighted version is, that it works better for long queries, distributing the weights as per the similarity values. 

For SIM Vec, we experimented with both the `Sum' and `Max' functions. After doing an analysis on the queries returned by the sum function, we found that those words that are related to the meaning of the entire query come up, while in max, words that have high similarity with one of the query terms, come up in the translation. Table \ref{tab:eg-queries-sum-max} illustrates some example queries from this method. For the first example, `sum' could not retrieve words like `assault' and `attack', because these were similar only to one query term, `{\sl hamalaa}', but not the others.



While the SIM Vec with the `Max' function performs the best among the proposed approaches, these results are still inferior to the monolingual baseline as well as Google Translate. Next, we use our proposed method with dictionary based approach as well as Google Translate in a hybrid model.

\subsection{Experiments with Hybrid Models}
For these experiments, we combine the dictionary based translations or those obtained from Google Translate with translations derived from the embedding based method. The following variations have been tried.

\begin{itemize}[leftmargin=*]
\item \textbf{Hybrid Translations using Dictionary (WE+DT)}: In this technique of query translation, for each query term q{$_i$}, we take translations from the dictionary, if a translation exists. If not, we take its translation from the embedding based methods. 
\item \textbf{Hybrid Translations using Dictionary, weighted (WE+DT weighted, SIM Vec+DT Weighted)}: We assign weights to the dictionary and word embedding based translation words such that the weights for the translations for each of the query terms add up to 1. If a query term has its translation from both dictionary as well as embedding based method, then the dictionary terms are assigned a total weight of $w$ and the rest ${1-w}$ is divided proportionately according to similarity values from the embedding based methods. We give 80$\%$ importance to the word embedding based terms and 20$\%$ importance to the dictionary based terms ($w=0.2$)\footnote{We experimented with other weightages like 70$\%$-30$\%$ and 90$\%$-10$\%$ but the 80$\%$-20$\%$ division gives the best result. We also experimented with the unweighted version of SIM Vec, but results were better with the weighted version and hence we omit them for brevity.} 
\item \textbf{Hybrid Translations using Google Translate\\ (Google Translate+Sim Vec, Google Translate+Sim Vec+DT)}: We include query translations from Google, with the same weighting approach as described above.
\end{itemize}

Table \ref{tab:we+dict} shows the results of the hybrid approaches with dictionary and Table \ref{tab:we+dic+gt} shows these results while using Google Translate with our embedding methods. In both the cases, the hybrid model improves upon the Dictionary / Google Traslate results, obtained when the word embeddings are not used. Specifically, Sim Vec with the Max function performs the best.

Results for some of the individual queries are shown in Table \ref{tab:eg-queries-dt-wt}. We see that WE, when combined with DT, retrieves many relevant terms, which improve the performance.
\begin{table}
\vspace{-0.2cm}
\small
    \centering
    \caption{Performance Results when Queries are Translated using a Hybrid of Word Embeddings and Dictionary}
    \label{tab:we+dict}
    \noindent\begin{tabular}{|c|c|c|c|c|c|c|c|}
    \hline
         & & \multicolumn{3}{c|}{\textbf{2012 Dataset}} & \multicolumn{3}{c|}{\textbf{2008 Dataset}}  \\
         \cline{3-8}
         \textbf{Method} & \textbf{\# Translations} & \textbf{MAP} & \textbf{P5} & \textbf{P10} & \textbf{MAP} & \textbf{P5} & \textbf{P10}\\
         \hline
         Dictionary	& - & 0.1691 & 0.2048	& 0.2048 & 0.0804 &	0.1464	& 0.137\\
         \hline
         \multirow{2}{*}{WE+DT} & 3words & {0.2593} & {0.404} & {0.38} & {0.128} & {0.168} & {0.16}\\
         
         & 5words & \textbf{0.2615} & \textbf{0.424} & \textbf{0.41} & \textbf{0.133} & \textbf{0.1835} & {0.168}\\
         
         & 7words & 0.26 & 0.416 & 0.397 & 0.13 & 0.174 & \textbf{0.169}\\
         \hline
          
         \multirow{2}{*}{WE+DT weighted} & 3words & {0.2623} & {0.358} & {0.35} & {0.1219} & {0.208} & {0.11}\\
         
         & 5words & \textbf{0.2898} & \textbf{0.4920} & \textbf{0.49} & \textbf{0.147} & \textbf{0.22} & \textbf{0.218}\\
         
         & 7words & 0.2718 & 0.391 & 0.39 & 0.136 & 0.19 & 0.18\\
         \hline
        
          \multirow{6}{*}{SIM Vec+ DT weighted}
         & Sum - 15words & 0.2835 & 0.4604 & 0.457 & 0.1419 & 0.237 & 0.23\\
         & Sum - 20words & \textbf{0.2850} & \textbf{0.4668} & \textbf{0.46} & \textbf{0.142} & \textbf{0.25} & \textbf{0.248}\\
         & Sum - 25words & 0.2824 & 0.4615 & 0.453 & 0.14 & 0.247 & 0.24\\
       
         \cline{2-8}
         & Max - 15words & 0.2965 & 0.495 & 0.49 & 0.148 & 0.234 & 0.228\\
         & Max - 20words & \textbf{0.2975} & \textbf{0.508} & \textbf{0.4913} & \textbf{0.1486} & 0.241 & 0.236\\
         & Max - 25words & 0.2967 & 0.497 & 0.485 & 0.139 & \textbf{0.25} & \textbf{0.248}\\
         
         \hline
           \end{tabular}
    \vspace{-0.2cm}
\end{table}
\begin{table}[h]
\vspace{-0.2cm}
\small
\centering
\caption{Example queries to illustrate the hybrid model with word Embeddings and Dictionary}
\label{tab:eg-queries-dt-wt}
\resizebox{\textwidth}{!}{%
\noindent\begin{tabular}{|c|c|c|c|c|c|c|}
\hline
\textbf{Query in Hindi} & \begin{tabular}[c]{@{}c@{}}\textbf{Translation in} \\ \textbf{English}\end{tabular} & \begin{tabular}[c]{@{}c@{}}\textbf{Translation}\\ \textbf{Method}\end{tabular} & \textbf{Translations} & \textbf{MAP} & \textbf{P5} & \textbf{P10} \\
\hline
\multirow{3}{*}{\begin{tabular}[c]{@{}c@{}}{\sl gorakhaalaiMDa} \\ {\sl kii maaMga}\end{tabular}} & \multirow{3}{*}{\begin{tabular}[c]{@{}c@{}}Demand of\\ Gorkhaland\end{tabular}} & DT & Gorkhaland & ~ 0.197 ~ & ~ 0.2 ~ & ~ 0.4 ~ \\
\cline{3-7}
 &  & \multirow{2}{*}{\begin{tabular}[c]{@{}c@{}}WE + DT\\ Weighted\end{tabular}} & \multirow{2}{*}{\begin{tabular}[c]{@{}c@{}}Gorkhaland\textasciicircum 1 demand\textasciicircum 0.51\\ demands\textasciicircum 0.49\end{tabular}} & \multirow{2}{*}{~ 0.88 ~} & \multirow{2}{*}{~ 1 ~} & \multirow{2}{*}{~1~} \\
 &  &  &  &  &  &  \\
 \hline
\multirow{2}{*}{\begin{tabular}[c]{@{}c@{}}{\sl abhiyukta} \\ {\sl ajamala}\\ {\sl kasaaba}\end{tabular}} & \multirow{2}{*}{\begin{tabular}[c]{@{}c@{}}Accused \\ Ajmal \\ Kasab\end{tabular}} & DT & Ajmal Kasab accused & ~ 0.32 ~ & ~ 0.2 ~ & ~ 0.2~ \\
\cline{3-7}
 &  & \begin{tabular}[c]{@{}c@{}}WE + DT \\ Weighted\end{tabular} & \begin{tabular}[c]{@{}c@{}}Ajmal\textasciicircum 1 Kasab\textasciicircum 1 murder\textasciicircum 0.26 \\ criminal\textasciicircum 0.25 murderer\textasciicircum 0.25 \\ complainant\textasciicircum 0.24 accusedˆ0.2\end{tabular} & ~ 0.66 ~ & ~ 0.8 ~ & ~ 0.8 ~ \\
 \hline
\multirow{2}{*}{\begin{tabular}[c]{@{}c@{}}2003 \\ {\sl aashiyaana}\\ {\sl kapa} \\ {\sl vijetaa}\end{tabular}} & \multirow{2}{*}{\begin{tabular}[c]{@{}c@{}}2003\\ ASEAN \\ Cup\\ winner\end{tabular}} & DT & \begin{tabular}[c]{@{}c@{}}2003 ASEAN cup \\ champion victor\end{tabular} & ~ 0.24 ~ & ~ 0.4 ~ & ~ 0.3 ~ \\
\cline{3-7}
 &  & \multicolumn{1}{l|}{\begin{tabular}[c]{@{}c@{}} ~ WE + DT ~ \\ Weighted\end{tabular}} & \begin{tabular}[c]{@{}c@{}}2003\textasciicircum 1 ASEAN\textasciicircum 1 \\ tournament\textasciicircum 0.8 cup\textasciicircum 0.2 \\ winners\textasciicircum 0.52  winner\textasciicircum 0.48\\ championship\textasciicircum 0.1 victor\textasciicircum 0.1\end{tabular} & ~ 0.4 ~ & ~ 0.6 ~ & ~ 0.5 ~\\
 \hline
\end{tabular}
}
\vspace{-0.27cm}
\end{table}
\begin{table}[!thb]
\vspace{-0.25cm}
\small
    \centering
    \caption{Performance Results when Queries are Translated using a Hybrid of Word Embeddings, Google Translate and Dictionary}
    \label{tab:we+dic+gt}
    \noindent\begin{tabular}{|c|c|c|c|c|c|c|c|}
    \hline
         & & \multicolumn{3}{c|}{\textbf{2012 Dataset}} & \multicolumn{3}{c|}{\textbf{2008 Dataset}}  \\
         \cline{3-8}
         \textbf{Method} & \textbf{\# Translations} & \textbf{MAP} & \textbf{P5} & \textbf{P10} & \textbf{MAP} & \textbf{P5} & \textbf{P10}\\
         \hline
          Google Translate & - & 0.3566 & 0.576 & 0.522 & 0.178 &	0.255 &	0.24\\
          \hline
         \multirow {6}{*}{Google Translate+Sim Vec} & 10words &	0.3669 & 0.552 & 0.532 & \textbf{0.184}	& 0.266 &	0.247\\
         & 13words & \textbf{0.3704} & 0.548 & 0.536 & 0.1798	& 0.278 & \textbf{0.249}\\
         & 15words &	0.3694 & 0.532 & 0.536	& 0.1737 & 0.271	& 0.243\\
         & 20words & 0.3691 &	0.552 &	\textbf{0.538} &	 0.173 & 0.276	& 0.235\\
         & 25words & 0.3699 & 0.568 & 0.532	&	0.1729	& \textbf{0.284} & 0.232\\
         & 30words & 0.3691 & \textbf{0.58}	& 0.53	& 0.1719 & 0.28	& 0.232\\
         \hline
         & 10words & 0.3682 & 0.556 & 0.526 & 0.1803 & 0.248  & 0.236  \\
Google Translate+         & 15words & \textbf{0.3719} & 0.56  & 0.532 & \textbf{0.1854} & 0.2506 & 0.2404 \\
Sim Vec+             & 20words & 0.3699 & 0.568 & 0.532 & 0.1776 & 0.253  & 0.2458 \\
          DT       & 25words & 0.3691 & 0.58  & 0.53  & 0.1727 & 0.2574 & \textbf{0.2492} \\
                 & 30words & 0.368  & \textbf{0.588} & \textbf{0.544} & 0.1703 & \textbf{0.2626} & 0.249  \\
                 \hline
         
    \end{tabular}
    
    \label{results}
    \vspace{-0.3cm}
\end{table}
 

From Table \ref{tab:we+dic+gt}, we see that our proposed method not only improves upon the dictionary but also improves over Google Translate and English Monolingual.
Table \ref{tab:perf-with-baselines} summarizes the improvements of our approach over the baselines, to nearest integers. For DT and \cite{clef:07}, improvements obtained by our method are shown, while for English Monolingual, we show the $\%$ of the E.M. results obtained by our method. We see that all the proposed approaches improve over DT and \cite{clef:07} consistently. Hybrid model with Google Translate improves even on the English monolingual.

\begin{table}[!thb]
\small
\centering
\caption{Comparison of Word Embedding based methods with Baselines. ( DT stands for `Dictionary' ; \cite{clef:07} refers to Chinnakotla et.al's method ; E.M. stands for `English Monolingual' ; imp. is `improvement' ) }
\label{tab:perf-with-baselines}
\begin{tabular}{|c|c|c|c|c|c|c|c|}
\hline
\multicolumn{2}{|c|}{\multirow{2}{*}{\textbf{Method}}} & \multicolumn{3}{c}{\textbf{2012 Dataset}} & \multicolumn{3}{|c|}{\textbf{2008 Dataset}} \\
\cline{3-8}
\multicolumn{2}{|c|}{} & {\begin{tabular}[c]{@{}c@{}} \textbf{$\%$ imp.}\\ \textbf{over DT}\end{tabular}} &  {\begin{tabular}[c]{@{}c@{}}\textbf{$\%$ imp.}\\ \textbf{over \cite{clef:07}} \end{tabular}} & \begin{tabular}[c]{@{}c@{}}\textbf{$\%$ of }\\ \textbf{E.M.}\end{tabular} & {\begin{tabular}[c]{@{}c@{}} \textbf{$\%$ imp.}\\ \textbf{over DT}\end{tabular}} &  {\begin{tabular}[c]{@{}c@{}}\textbf{$\%$ imp.}\\ \textbf{over \cite{clef:07}} \end{tabular}} & \begin{tabular}[c]{@{}c@{}}\textbf{$\%$ of }\\ \textbf{E.M.}\end{tabular} \\
\hline
\multirow{4}{*}{Simple} & WE & ~ 52 ~ & ~ 15 ~ & ~ 80 ~ & ~ 54 ~ & ~ 17 ~ & ~ 80 ~ \\
\cline{2-8}
 & WE weighted & ~ 66 ~ & ~ 26 ~ & ~ 87 ~ & ~ 66 ~ & ~ 27 ~ & ~ 86 ~ \\
 \cline{2-8}
 & \begin{tabular}[c]{@{}c@{}}SIM Vec - sum\end{tabular} & ~ 52 ~ & ~ 15 ~ & ~ 80 ~ & ~ 53 ~ & ~ 17 ~ & ~ 80 ~ \\
 \cline{2-8}
 & \begin{tabular}[c]{@{}c@{}}SIM Vec - max\end{tabular} & \textbf{~ 68 ~ } & \textbf{~ 27 ~} & \textbf{~ 88 ~} & \textbf{~ 72 ~} & \textbf{~ 31 ~} & \textbf{~ 89 ~} \\
 \hline
\multirow{4}{*}{\begin{tabular}[c]{@{}c@{}}Hybrid\\ with\\ Dictionary\end{tabular}} & WE + DT & ~ 55 ~ & ~ 17 ~ & ~ 81 ~ & ~ 58 ~ & ~ 21 ~ & ~ 83 ~ \\
\cline{2-8}
 & \begin{tabular}[c]{@{}c@{}}WE + DT weighted\end{tabular} & ~ 72 ~ & ~ 30 ~ & ~ 90 ~ & ~ 75 ~ & ~ 34 ~ & ~ 91 ~ \\
 \cline{2-8}
 & \begin{tabular}[c]{@{}c@{}}SIM Vec (sum)+\\ DT\end{tabular} & ~ 69 ~ & ~ 27 ~ & ~ 89 ~ & ~ 69 ~ & ~ 29 ~ & ~ 89 ~ \\
 \cline{2-8}
 & \begin{tabular}[c]{@{}c@{}}SIM Vec (max)+\\ DT\end{tabular} & \textbf{~ 76 ~} & \textbf{~ 33 ~} & \textbf{~ 92 ~} & \textbf{~ 77 ~} & \textbf{~ 35 ~} & \textbf{~ 92 ~} \\
 \hline
\multirow{2}{*}{\begin{tabular}[c]{@{}c@{}}Hybrid with\\ Dictionary\\ and Google\\ Translate\end{tabular}} & \begin{tabular}[c]{@{}c@{}}GT + SIM \\ Vec (max)\end{tabular} & ~ 119 ~ & ~ 66 ~ & ~ 115 ~ & ~ 119 ~ & ~ 66 ~ & ~ 114 ~ \\
\cline{2-8}
 & \begin{tabular}[c]{@{}c@{}}GT + SIM\\ Vec (max) +\\ DT\end{tabular} & ~ \textbf{119} ~ & ~ \textbf{66} ~ & ~ \textbf{115} ~ & ~ \textbf{120} ~ & ~ \textbf{69} ~ & ~ \textbf{115} ~\\
 \hline
\end{tabular}
 \vspace{-0.6cm}
\end{table}

\section{Conclusion and Future Work}
In this paper, we proposed a method based on word embeddings for query translation in the CLIR task. Extensive evaluations performed under various settings confirm that word embedding based method is a potential tool with which the language barrier in the CLIR task can be resolved. It alone performs well over the dictionary method and when combined with the dictionary and Google Translate in a hybrid model, it gives the best performance, improving even the target monolingual baseline by 15$\%$. 
In future, we will like to repeat these experiments over other source-target language pairs to confirm that this is generalizable across many different language pairs and achieves similar performance gains. We will also study the effect of corpus size (on source and target) as well as the dictionary size on the performance of the system. Finally, we will also experiment using this method for tasks such as bilingual lexical induction.
\section{Acknowledgments}
We would like to thank the anonymous reviewers for their valuable comments. This work is supported by the project "To Develop a Scientific Rationale of IELS
(Indo-European Language Systems) Applying A) Computational Linguistics
\& B) Cognitive Geo-Spatial Mapping Approaches" funded by the Ministry of Human Resource Development (MHRD), India and conducted in Artificial Intelligence Laboratory, Indian Institute of Technology Kharagpur.
\bibliography{biblio}
\bibliographystyle{splncs03}
\end{document}